\pdfoutput=1

\documentclass[11pt]{article}

\usepackage[preprint]{acl}

\usepackage{times}
\usepackage{latexsym}

\usepackage[T1]{fontenc}

\usepackage[utf8]{inputenc}

\usepackage{microtype}

\usepackage{inconsolata}

\usepackage{graphicx}

\usepackage{adjustbox}
\usepackage{multirow}
\usepackage{booktabs} 
\usepackage{amssymb}
\usepackage{amsmath}
\usepackage{subfigure}
\usepackage{makecell}
\usepackage{color}
\usepackage[normalem]{ulem}
\newcolumntype{Y}{>{\centering\arraybackslash}X}
\newcolumntype{C}[1]{>{\centering\arraybackslash}p{#1}}

\usepackage{enumitem}
\setlist[itemize]{noitemsep, topsep=1pt}

\title{MoDE: Effective Multi-task Parameter Efficient Fine-Tuning with a Mixture of Dyadic Experts}

\author{
\setcounter{footnote}{1}
 \textbf{Lin Ning\footnotemark[1]}, 
 \textbf{Harsh Lara}, 
 \textbf{Meiqi Guo}, 
 \textbf{Abhinav Rastogi\footnotemark[1]\footnotemark[2]}
\\
\\ 
  Google DeepMind
\\
\texttt{\{linning, harshlara, mqguo, abhirast\}@google.com}
}

\begin{document}

\maketitle
\footnotetext[1]{Equal contribution.}
\footnotetext[2]{Corresponding author.}

\begin{abstract}
Parameter-efficient fine-tuning techniques like Low-Rank Adaptation (LoRA) have revolutionized the adaptation of large language models (LLMs) to diverse tasks. Recent efforts have explored mixtures of LoRA modules for multi-task settings. However, our analysis reveals redundancy in the down-projection matrices of these architectures. This observation motivates our proposed method, \textbf{M}ixture \textbf{o}f \textbf{D}yadic \textbf{E}xperts \textbf{(MoDE)}, which introduces a novel design for efficient multi-task adaptation. This is done by sharing the down-projection matrix across tasks and employing atomic rank-one adapters, coupled with routers that allow more sophisticated task-level specialization. Our design allows for more fine-grained mixing, thereby increasing the model's ability to jointly handle multiple tasks. We evaluate MoDE on the Supernatural Instructions (SNI) benchmark consisting of a diverse set of 700+ tasks and demonstrate that it outperforms state-of-the-art multi-task parameter-efficient fine-tuning (PEFT) methods, without introducing additional parameters. Our findings contribute to a deeper understanding of parameter efficiency in multi-task LLM adaptation and provide a practical solution for deploying high-performing, lightweight models.
\end{abstract}

\section{Introduction}
\label{sec:introduction}

\begin{figure}[ht]
\centering
    \includegraphics[width=\columnwidth]{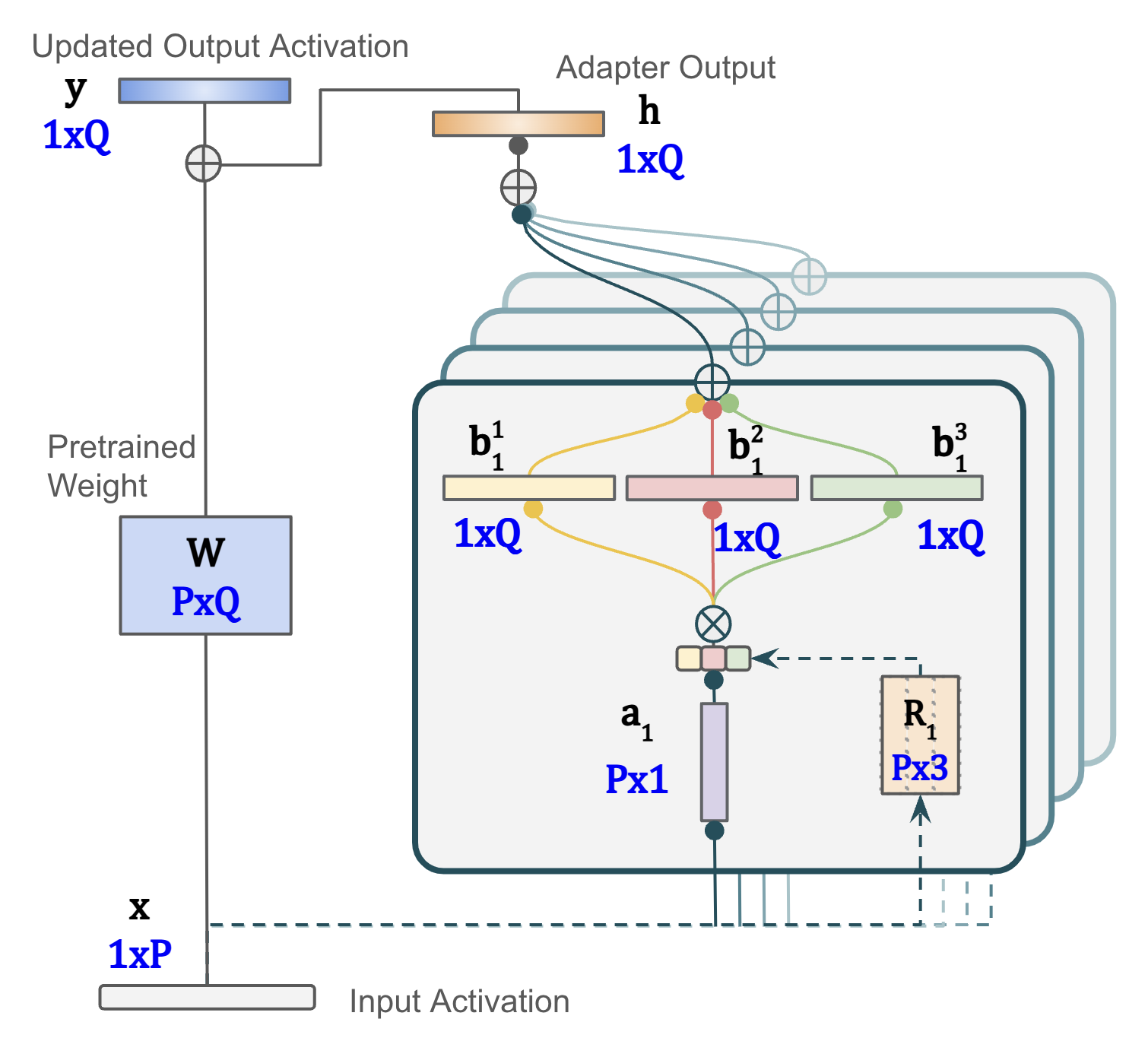}
    \vspace{-0.2in}
    \caption{Mixture of Dyadic Experts with 3 experts and a rank of 4, with each slice corresponding to a rank dimension. Our architecture allows independent routing at each rank. When number of mixtures is 1, our architecture is equivalent to traditional LoRA.}
    \vskip -0.2in
    \label{fig:MoRA}
\end{figure}

Large language models (LLMs) have demonstrated remarkable capabilities in various natural language processing tasks, from text generation and translation to question-answering and summarization \cite{brown2020language, team2023gemini, team2024gemma, openai2024gpt4technicalreport}. The ability to perform well on a diverse set of tasks is essential for deploying LLMs in real-world applications, where they may need to handle a wide array of user requests and instructions. However, effectively adapting these large models to multiple tasks presents significant challenges. Fine-tuning a separate model for each task is computationally expensive and requires vast amount of storage due to the  large model sizes. Moreover, independently trained models hinder knowledge transfer between tasks, potentially limiting the model's performance and its ability to generalize to unseen tasks.

Multi-task learning (MTL) \cite{caruana1997multitask, ruder2017overview} offers a promising solution to these challenges. By training a single model on multiple tasks simultaneously, MTL aims to improve parameter efficiency, enhance generalization, and potentially boost performance on individual tasks through knowledge transfer. Parameter-efficient fine-tuning (PEFT) techniques, such as Low-Rank Adaptation (LoRA) \cite{hu2021lora}, have further enhanced efficiency by introducing only a small number of trainable parameters. LoRA efficiently represents weight changes during fine-tuning using two low-rank projection matrices: one down-projects input features to a smaller size, and another up-projects the resulting low-dimensional representation back to the original output size. 

Mixture-of-Experts (MoE) architectures \cite{sukhbaatar2024branchtrainmixmixingexpertllms,li2022branchtrainmergeembarrassinglyparalleltraining,jiang2024mixtralexperts,fedus2022switchtransformersscalingtrillion} have emerged as a powerful approach to scale model capacity and expertise, enabling LLMs to handle a wider range of tasks. Recent studies \cite{feng2024mixture, zhu2023sira, zadouri2023pushing, liu2023moelora,li2024mixlora} have explored integrating LoRA with Mixture-of-Experts architectures (LoRA-MoE) to extend LLMs' capabilities to multi-task adaptation. However, our analysis reveals redundancy in down-projection matrices of these architectures. This redundancy leads to an inefficient utilization of parameters, potentially limiting the effectiveness in capturing the unique characteristics of each task.

In this work, we propose Mixture of Dyadic Experts (MoDE) (Figure \ref{fig:MoRA}), a novel parameter-efficient framework for multi-task adaptation. MoDE leverages a single, shared down-projection matrix across all experts to reduce parameter redundancy. Furthermore, MoDE introduces atomic rank-one adapters, enabling fine-grained task specialization and knowledge sharing. Crucially, MoDE incorporates a sophisticated routing mechanism that allows for more nuanced and flexible combinations of these rank-one adapters, further enhancing the model's expressive power while maintaining parameter efficiency.

We rigorously evaluate MoDE on the multi-task Supernatural Instructions benchmark \cite{wang-2022-sni}. Our results demonstrate that MoDE consistently outperforms state-of-the-art multi-task PEFT methods, including those based on LoRA-MoE, while utilizing comparable number of additional parameters. This underscores MoDE's efficacy in achieving both strong performance and parameter efficiency, making it a promising approach for deploying multi-task LLMs in real-world applications.

Our \textbf{key contributions} are as follows:
\begin{itemize}
\setlength\itemsep{0.3em}
    \item We identify and address the redundancy in down-projection matrices in existing LoRA-based MoE architectures. 
    \item We propose MoDE, a novel architecture leveraging a shared down-projection matrix and atomic rank-one adapters, coupled with a sophisticated routing mechanism for efficient and expressive performance.
    \item We demonstrate the superior performance of MoDE compared to state-of-the-art multi-task LoRA-based MoE methods on the Supernatural Instructions benchmark, while maintaining parameter efficiency.
\end{itemize}

\section{Related Work}
\label{sec:related}

\subsection{Parameter-efficient Fine-tuning (PEFT)}
Parameter-efficient fine-tuning (PEFT) methods have emerged as a popular approach to adapt LLMs to downstream tasks  without the high computational cost of full fine-tuning. LoRA (Low-Rank Adaptation) \cite{hu2021lora} is a particularly successful PEFT technique, achieving strong performance with a fraction of the trainable parameters.

\subsection{Mixture-of-Experts and LoRA}
Recent research has explored combining Mixture-of-Experts (MoE) architectures with LoRA to further improve efficiency and effectiveness, especially in multi-task scenarios. One notable approach is to propose frameworks that leverage domain-specific LoRA modules and explicit routing strategies to adapt to diverse tasks, such as Mixture-of-LoRAs (MoA) \cite{feng2024mixture} and MOELoRA \cite{liu2023moelora}. Another group of work, represented by SiRA \cite{zhu2023sira} and MixLoRA \cite{li2024mixlora}, introduce sparse MoE mechanisms with specialized routing and/or load-balancing techniques to enhance efficiency while maintaining performance. MoLORA \cite{zadouri2023pushing} combines MoE with LoRA experts to achieve extreme parameter efficiency in instruction tuning. Their approach focuses on updating only a small fraction of the model's parameters, demonstrating comparable performance to full fine-tuning with significantly fewer resources. AdaMix \cite{wang2022adamix} proposes a general PEFT method that tunes a mixture of adaptation modules within each Transformer layer to capture multiple views of a single task, using LoRA module sharing to enhance performance when labeled data is scarce. Their experiment show that sharing project-up has better performance. In contrast, our multi-task approach leverages a mixture of LoRA modules with shared project-down matrices, motivated by the observation of their similarity across different tasks.

\subsection{Multi-task PEFT}
Other work has focused on extending PEFT to multi-task settings, where a single model needs to adapt to diverse tasks.
LoraHub \cite{huang2023lorahub} investigates LoRA composability for cross-task generalization and introduces a framework for dynamically assembling LoRA modules trained on different tasks to adapt to unseen tasks.
ZipLoRA \cite{shah2023ziplora} tackles the problem of combining independently trained style and subject LoRAs to achieve joint generation in a controllable manner.
FLix \cite{lin2024multitask} focuses on multi-task multilingual model adaptation, associating each dataset feature with its own low-rank weight update parameters for improved generalization across diverse datasets.
MoLE \cite{wu2024mixture}  implements a hierarchical weight control approach with learnable gating functions to determine the optimal composition of trained LoRA layers, treating each layer as a distinct expert. 
Different from composing over all LoRA adaptors, \citet{ostapenko2024towards} explores building a library of trained LoRA adapters and using a zero-shot routing mechanism (Arrow) to dynamically select relevant adapters for new tasks.


\section{Method}
\label{sec:mora}


The Mixture of Dyadic Experts (MoDE) architecture presents a novel approach for multi-task learning, building upon and extending the traditional Low-Rank Adaptation (LoRA) and mixture-of-experts (MoE) design. 

\subsection{Background}
\label{subsec:background}

\begin{figure}[t]
\centering
    \includegraphics[width=0.5\columnwidth]{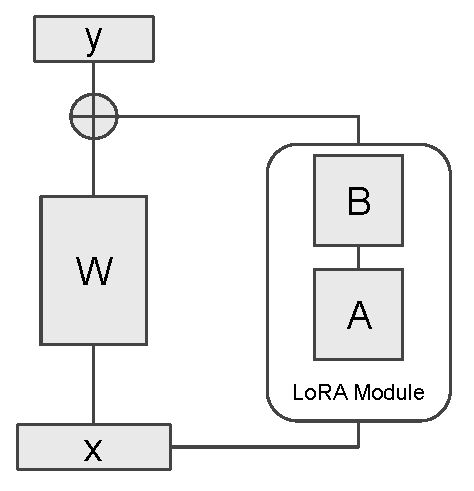}
    \vskip -0.1in
    \caption{Illustration of a basic LoRA module.}
    \vskip -0.1in
    \label{fig:LoRA}
\end{figure}

\paragraph{Low-Rank Adaption (LoRA)} LoRA \cite{hu2021lora} efficiently adapts LLMs to downstream tasks \cite{shah2023ziplora} by freezing pre-trained model weights and injecting trainable rank decomposition matrices into each layer. Given a feed-forward layer with input $\mathbf{x} \in \mathbb{R}^{1\times P}$ and weight matrix $\mathbf{W_0} \in \mathbb{R}^{P\times Q}$, LoRA introduces a down-projection matrix $\mathbf{A} \in \mathbb{R}^{P\times r}$ and an up-projection matrix $\mathbf{B} \in \mathbb{R}^{Q\times r}$ (Figure \ref{fig:LoRA}). The output of the layer is
$$\mathbf{y} = \mathbf{x}\mathbf{W_0} + \mathbf{x}\mathbf{A}\mathbf{B}^T.$$
During training, only $\mathbf{A}$ and $\mathbf{B}$ are updated.

\begin{figure}[t]
\centering
    \subfigure[]{\includegraphics[width=0.23\textwidth]{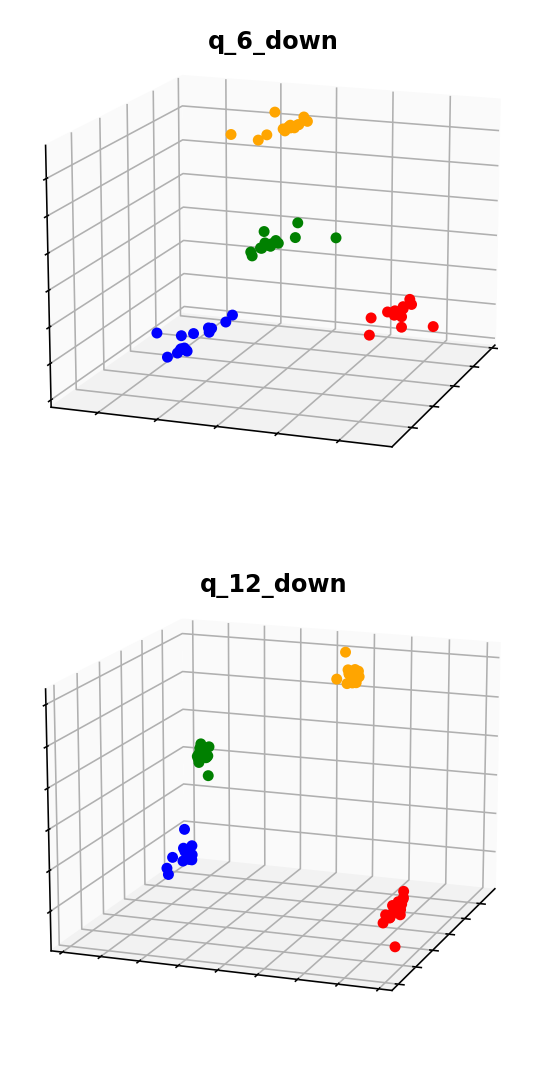}}
    \subfigure[]{\includegraphics[width=0.23\textwidth]{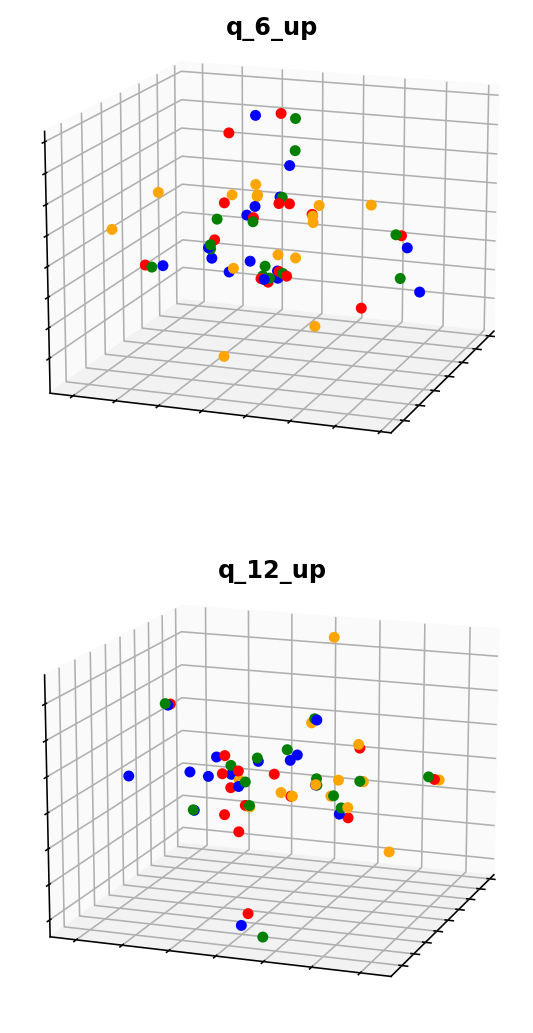}}
    \vskip -0.1in
    \caption{Scatter plots showing the three most prominent principal components of all constituent vectors in the LoRA projection matrices for 15 independently trained single task model with shared initialization. Plots q\_6\_down and q\_12\_down (q\_6\_up and q\_12\_up) illustrate the down (up) projections of query matrices at layers 6 and 12, respectively. The clear clustering of down-projection vectors suggests that the down projection matrices are task-agnostic, motivating the design of the MoDE architecture.}
    \vskip -0.1in
    \label{fig:PCA_plots}
\end{figure}

\begin{figure*}[ht]
\centering
    \subfigure[]{\includegraphics[width=0.48\textwidth]{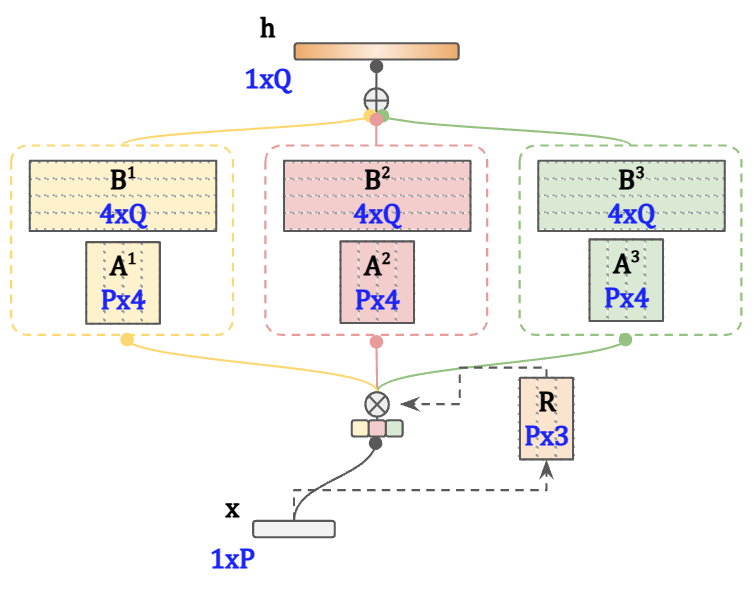}}
    \subfigure[]{\includegraphics[width=0.48\textwidth]{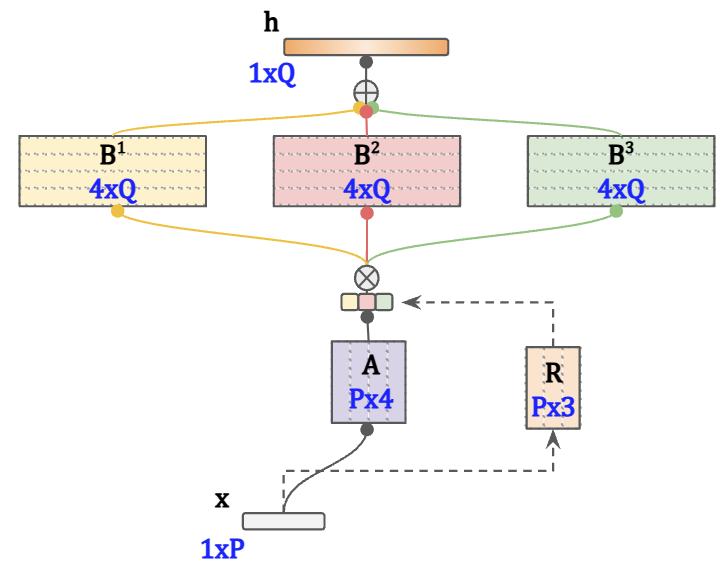}}
    \vskip -0.1in
    \caption{Illustration of (a) traditional LoRA Mixture-of-Experts, (b) traditional LoRA Mixture-of-Experts with shared down-projection matrix.}
    \vskip -0.1in
    \label{fig:SiRA}
\end{figure*}

\paragraph{Dyadic Product Representation}
A dyadic product (or outer product) is a matrix multiplication between two vectors. Given vectors $\mathbf{u} \in \mathbb{R}^{p \times 1}$ and $\mathbf{v} \in \mathbb{R}^{q \times 1}$, their dyadic product $\mathbf{u} \otimes \mathbf{v}$ is a matrix of size $p \times q$. The LoRA update $\Delta \mathbf{W} = \mathbf{A}\mathbf{B}^T$ can be expressed as a sum of dyadic products by decomposing $\mathbf{A}$ and $\mathbf{B}$ into their column vectors \cite{liu2024intuitionaware}:
\begin{align*}
\Delta \mathbf{W} &= [\mathbf{a}_1, \mathbf{a}_2, ..., \mathbf{a}_r] * [\mathbf{b}_1, \mathbf{b}_2, ..., \mathbf{b}_r]^T \\
&= \sum^r_{i=1}(\mathbf{a}_i \otimes \mathbf{b}_i)
\end{align*}
where $\mathbf{a}_i$ and $\mathbf{b}_i$ are column vectors of $\mathbf{A}$ and $\mathbf{B}$, respectively. This can be plugged into the output equation to get:
$$\mathbf{y} = \mathbf{x}\mathbf{W_0} + \mathbf{x}\sum^r_{i=1}(\mathbf{a}_i \otimes \mathbf{b}_i)$$

\paragraph{LoRA-MoE}
Mixture-of-Experts (MoE) utilizes a combination of sub-models (experts), each specializing in different aspects of the underlying tasks, along with a gating mechanism to dynamically route inputs to the most suitable experts \cite{shazeer2017outrageously}. Figure \ref{fig:SiRA})(a) illustrates a traditional LoRA-MoE approach \cite{zadouri2023pushing}, where $m$ LoRA experts ($E^i = \mathbf{A}^i {\mathbf{B}^i}^T$, $i \in \{1, ..., m\}$) are added to each layer. A router $\mathcal{R}$, parameterized by $\mathbf{W}_R \in \mathbb{R}^{P \times m}$, determines which expert to use, yielding the output:
$$
\mathbf{y} = \mathbf{x}\mathbf{W_0} + \sum_{i=1}^m\mathcal{R}^i(\mathbf{x})(\mathbf{x}\mathbf{A}^i{\mathbf{B}^i}^T)
$$
where $\mathcal{R}^i(\mathbf{x})$ is the routing probability for $E^i$.

\subsection{Motivating Observation}
\label{subsec:single_task_lora_clustering}
To motivate our proposed MoDE architecture, we first present an empirical analysis of projection matrices of LoRA modules that are independently trained on a set of tasks from the same initialization ($B_i$ to $0$ and $A_i$ from a random normal distribution with a standard deviation of $0.01$ and mean $0$). We selected 15 diverse tasks from the Supernatural Instructions benchmark (see Section \ref{subsec:dataset-and-metrics} for details) and trained 15 LoRA modules for this study. We visualized the learned LoRA parameters using Principal Component Analysis (PCA), focusing on the distribution of vectors obtained by slicing the up-projection and down-projection matrices along their rank dimension, i.e., the vectors featuring in the dyadic product representation in Section \ref{subsec:background}.

Figure \ref{fig:PCA_plots} shows the resulting scatter plots. Notably, we observe that down-projection matrix vectors from different LoRA modules tend to cluster into distinct groups, with vectors corresponding to the same position along the rank dimension forming tight clusters. In contrast, up-projection matrices exhibit no such clustering. This suggests down-projection matrices are task-agnostic, while up-projection matrices are more task-specific.

This empirical finding motivates the MoDE architecture, which leverages a shared down-projection matrix to reduce parameter redundancy. We further improve this design by leveraging the dyadic formulation to introduce a more sophisticated routing strategy which enables a more fine-grained task-specific adaptation. Subsequent sections will detail MoDE's architecture and demonstrate its effectiveness in achieving both parameter efficiency and strong multi-task performance.

\subsection{Mixture of Dyadic Experts (MoDE)}
Inspired by the observations in Section \ref{subsec:single_task_lora_clustering}, we introduce Mixture of Dyadic Experts (MoDE), a novel framework for efficient multi-task adaptation that incorporates two key innovations: (i) shared down-projection matrices for more efficient parameter utilization, and (ii) a sophisticated routing strategy to promote better task-level specialization.

\subsubsection{Shared Down-Projection Matrix}
Before introducing MoDE, we share a simple modification of the traditional LoRA-MoE by having all experts share a single down-projection matrix $\mathbf{A}$ (Figure \ref{fig:SiRA}(b)), which we refer to as LoRA-MoE-SD. The output of a layer with LoRA-MoE-SD is:
$$
\mathbf{y} = \mathbf{x}\mathbf{W_0} + \sum_{i=1}^m\mathcal{R}^i(\mathbf{x})(\mathbf{x}\mathbf{A}{\mathbf{B}^i}^T)
$$
where $\mathbf{A}$ is the shared down-projection matrix, and $\mathbf{B}^i$ is the up-projection matrix for expert $E_i$.

\paragraph{Parameter Efficiency}
By sharing a single down-projection matrix $\mathbf{A}$ across all experts, LoRA-MoE-SD reduces the number of trainable parameters for these matrices from $m\cdot P \cdot r$ to $P \cdot r$. 

\subsubsection{Fine-Grained Routing}
LoRA-MoE-SD, albeit with smaller parameters, offers only $m$ choices for up projection. This is because the router inherently introduces a constraint that all the $r$ dimensions of each expert must route together. In other words, the dyadic representation of LoRA-MoE update contains $m \times r$ terms, but the router only provides $m$ weights. MoDE addresses this by introducing atomic rank-one adapters. This design choice allows MoDE to leverage the dyadic product representation of LoRA, where each rank-one update captures a specific direction of change in the original weight matrix.


MoDE employs $m$ rank-one experts for each column vector $\mathbf{b_j}$ in $\mathbf{B}$ (Figure \ref{fig:MoRA}), resulting in $m \times r$ experts. Each expert $E_j^i$ (where $i \in \{1, ..., m\}$ and $j \in \{1, ... r\}$) specializes in a specific component of the up-projection, represented as a dyadic product $\mathbf{a}_j \otimes {\mathbf{b}^i_j}^T$, where $\mathbf{a}_j$ is the $j$-th column vector of the shared down-projection matrix $\mathbf{A}$ and $\mathbf{b}^i_{j}$ is the vector representing the $i$-th rank-one expert for the $j$-th component of $\mathbf{B}$.

Let $\mathcal{R}^i_j(\mathbf{x})$ is the routing probability for expert $E^i_j$ given input $\mathbf{x}$, the output with the MoDE module is:
$$
\mathbf{y} = \mathbf{x}\mathbf{W_0} + \sum_{i=1}^m\sum_{j=1}^r\mathcal{R}^i_j(\mathbf{x})(\mathbf{x} (\mathbf{a}_j \otimes {\mathbf{b}^i_j}^T)).
$$

\paragraph{Model Expressivity} The router in MoDE independently selects the expert used for each vector $\mathbf{b}_j$. This fine-grained control allows for flexible combination of these dyadic product experts, enabling MoDE to dynamically compose a specialized up-projection matrix tailored to the input and task. For example, if $\mathbf{B}$ has rank 4, the router might select $E_1^1$ for $\mathbf{b}_1$, $E_2^3$ for $\mathbf{b}_2$, $E_3^2$ for $\mathbf{b}_3$, and $E_4^1$ for $\mathbf{b}_4$.

With $m$ rank-one experts per vector $\mathbf{b}_j$ of a rank-$r$ up-projection matrix, MoDE can model $m^r$ different expert compositions, allowing for a wide range of task-specific adaptations compared to the $m$ experts in a traditional LoRA-MoE, given a similar number of parameters. This increased expressivity, derived from the flexible combination of dyadic products, allows MoDE to better capture the nuances of individual tasks while maintaining scalability for a large number of tasks.

\paragraph{MoDE Routing}
MoDE utilizes a token-level soft routing strategy, where the router $\mathcal{R}$ assign a weight to each rank-one expert for a given input token. The weighted sum of experts outputs determines the final output. This approach enables dynamic utilization of the most relevant experts for each input, facilitating nuanced and context-aware adaptation. 

The router network is denoted as $\mathbf{W}_\mathcal{R} \in \mathbb{R}^{r \times P\times m}$, where $\mathbf{W}_{\mathcal{R};j} \in \mathbb{R}^{P \times m}$ represents the network for vector $\mathbf{b}_j$ in the up-projection matrix. For an input $\mathbf{x}$, the routing weights $\mathcal{R}_j \in \mathbb{R}^{1\times m}$ for the experts corresponding to $\mathbf{b}_j$ are calculated as
$$\mathcal{R}_j(\mathbf{x}) = softmax(\mathbf{x} \cdot \mathbf{W}_{\mathcal{R};j}).$$
This mechanism allows MoDE to adaptively combine the expertise of multiple rank-one adapters, leading to improved multi-task performance.

\subsection{Generalization}
\label{sec:MoDE-generalization}
The rank-1 adapters in MoDE can be generalized to rank-$p$, where the router selects a composition of rank-$p$ adapters for each input. This requires the LoRA rank $r$ to be divisible by adapter rank $p$. The generalized output calculation becomes:

\begin{align*}
\mathbf{y} = \mathbf{x}\mathbf{W_0} + \sum_{i=1}^m\sum_{k=1}^{r/p}\mathcal{R}^i_k(\mathbf{x}) \cdot  \mathbf{x} \mathbf{A}_k{\mathbf{B}_k^i}^T
\end{align*}
where
\begin{align*}
\mathbf{A}_k{\mathbf{B}_k^i}^T = \sum_{j=1}^p (\mathbf{a}_{j+p(k-1)} \otimes {\mathbf{b}^i_{j+p(k-1)}}^T).  
\end{align*}

We denoted this generalized module as MoDE $m{\times}r{\times}p$. Note that MoDE $1{\times}r{\times}r$ is computationally equivalent to a LoRA module of rank $r$, and MoDE $m{\times}r{\times}r$ is computationally equivalent to LoRA-MoE-SD with rank $r$ and $m$ experts.
\section{Experiments}
\label{sec:experiments}

We comprehensively evaluate MoDE's performance and analyze its design choices through three sets of experiments on the Supernatural Instructions (SNI) benchmark \cite{wang-2022-sni}: (1) multi-task evaluation on the full dataset, (2) an ablation study on generalized MoDE architecture (Section \ref{sec:MoDE-generalization}) on the full dataset , and (3) a case study with a fixed number of tasks and parameter budgets. This section details the experimental setup and presents the results.

\subsection{Datasets and Metrics}
\label{subsec:dataset-and-metrics}
We leverage the Supernatural Instructions (SNI) dataset (with 1,616 diverse instruction-following tasks covering 76 distinct task types) for our experiments, focusing on the 756 English-only tasks from the default train split for both training and evaluation. For each task, we split the examples into a 90\% training set and 10\% evaluation set. For multi-task experiments, we create mixed training and evaluation datasets by combining examples from all 756 tasks.

We also curate a diverse subset of 15 individual tasks from the 756 tasks for the case study given a fixed parameter budgets. These tasks are selected from 15 different categories to ensure a comprehensive evaluation across different domains. Statistics about the sequence length of each task is shown in Tab.~\ref{tab:data_stats}. Each of the selected tasks contains more than 5k instances for training and from 500 to 650 instances for evaluation.

\begin{table}[]
\centering
\small
\begin{tabular}{p{2.9cm}C{0.7cm}cc} 
\toprule
Category (Task ID)  & Instruct & Input & Output\\ 
\midrule
\scriptsize{QuestionAnswering(24)} & 104 & 87/90.9 & 8/8.8   \\
\scriptsize{WrongCandidateGeneration(25)} & 127 & 100/104.1 & 8/8.8 \\
\scriptsize{QuestionGeneration(74)} & 97 & 143/155 & 12/12.5  \\
\scriptsize{GrammarErrorDetection(89)} & 89 & 10/10.6 & 6/6  \\
\scriptsize{LinguisticProbing(114)} & 66 & 25/25.7 & 1/1  \\
\scriptsize{PosTagging(155)}	& 19 & 23/23.7 & 1/1 \\
\scriptsize{Explanation(192)} & 43 & 123/131.4 & 30/33.7  \\
\scriptsize{StoryComposition(269)} & 184 & 82/82.2 & 34/4.7 \\ 
\scriptsize{StereotypeDetection(279)} & 89 & 15/15.5 & 2/2\\
\scriptsize{CommonsenseClassification(291)} & 54 & 17/17.8 & 1/1 \\
\scriptsize{ProgramExecution(622)} & 62 & 94/93.6 & 99/98.5   \\
\scriptsize{FillInTheBlank(672)} & 24 & 13/3.2 & 1/1   \\
\scriptsize{PoemGeneration(1711)}	& 83 & 3/3.6 & 44/59 \\
\scriptsize{DialogueGeneration(1729)} & 58 & 155/156.9 & 13/12.7  \\
\bottomrule
\end{tabular}
\caption{Sequence lengths of instruction, input (median/mean) and output (median/mean) in the selected SNI datasets for the fixed parameter budget case study.}
\vskip -0.1in
\label{tab:data_stats}
\end{table}

\paragraph{Evaluation Metric} We report ROUGE-L, the default metric for SNI dataset, for all experiments. 

\begin{table}[h]
\small
\centering
\begin{tabular}{l|cc}
\toprule
& Eval &  Add. Params. \\
\midrule
LoRA 64 &  56.11 & 6.31\% \\  
MoLORA 16$\times$4 & 57.77 & 7.62\% \\  
MoLORA-SD 16$\times$4 & 58.28 & 2.71\% \\
MoDE 16$\times$4 & {\bf 60.00} & 6.64\% \\   
\midrule
\midrule
MoDE 8$\times$4 & 59.00 & 3.48\% \\
MoDE 6$\times$4 & 60.91 & 2.69\% \\
MoDE 4$\times$4 & 60.18 & 1.90\% \\
MoDE 4$\times$6 & 60.53 & 2.86\% \\  
MoDE 4$\times$8 & 58.92 & 3.81\% \\ 
MoDE 4$\times$16 & 60.04 & 7.62\% \\  
\bottomrule
\end{tabular}
\caption{Row 2-5: multitask performance comparison between LoRA, MoLORA, MoLORA-SD and MoDE. Row 6-11: ablation study on MoDE with varying number of ranks and experts. The evaluation metric used is ROUGE-L. The last column represents the total number of adapter parameters as a percentage of the total number of non-embedding parameters in Gemma-2B.}
\vskip -0.2in
\label{tab:full-dataset-results}
\end{table}

\subsection{Implementation Details}
\paragraph{Model} The Gemma 2B language model \cite{team2024gemma} serves as the foundational LLM for all experiments due to its state-of-the-art performance on a variety of natural language processing tasks and its efficient size. 

\paragraph{Fine-tuning Setup} For all experiments, we fine-tune the parameter-efficient adaptors using Adafactor optimizer\cite{shazeer2018adafactor} with a learning rate of 1e-3, a total sequence length of 1024, and a batch size of 128 over 20,000 steps.

\subsection{Multi-task performance}
\label{subsec: multitask_results}

We assess model performance on the the full dataset comprising 756 tasks. We compare MoDE with vanilla LoRA and MoLORA \cite{zadouri2023pushing}, a strong baseline approach of LoRA-MoE for multi-task adaptation. To better understand the benefit of removing redundancy in down-projection matrices, we also apply down-projection sharing to MoLORA, referred to as MoLORA-SD. In addition to evaluating overall performance, we conduct an ablation study to investigate the impact of varying the number of experts ($m$) and the rank ($r$) of the LoRA matrices on MoDE's performances. 

The specific models and configurations we evaluate are as follows:
\begin{itemize}
  \item[$\circ$] {\em LoRA 64}: A base LoRA module with rank 64, having roughly the same number of parameters as other MoE models.
  \item[$\circ$] {\em MoLORA 16$\times$4}: A LoRA-MoE model with 4 experts, each using a rank-4 LoRA module.
  \item[$\circ$] {\em MoLORA-SD 16$\times$4}: A LoRA-MoE model with a shared down-projection matrix and 4 experts with rank-4 up-projection matrices.
  \item[$\circ$] {\em MoDE 16$\times$4}: A MoDE model with 4 experts per vector of a rank-4 up-projection matrix.
  \item[$\circ$] {\em MoDE m$\times$r}: MoDE models with different number of experts ($m$) and ranks ($r$) for ablation study. We explore the combinations 8$\times$4, 6$\times$4, 4$\times$4, 4$\times$6, 4$\times$8, and 4$\times$16.
\end{itemize} 

\begin{table}[t]
\small
\begin{tabular}{l|ccc}
\toprule
Model & LoRA & MoLORA & MoLORA-SD \\
\midrule
LoRA & \textbackslash{} & \textbackslash{} & \textbackslash{} \\
MoLORA  & 69\% & \textbackslash{} & \textbackslash{} \\
MoLORA-SD & 69\% & 60\% & \textbackslash{} \\
MoDE & 78\% & 73\% & 68\% \\
\bottomrule
\end{tabular}
\caption{Win rate against baseline models at the task-level evaluation.}
\vskip -0.2in
\label{tab:full-dataset-results-per-task}
\end{table}

\paragraph{Model Comparison}
Raw 2 - 5 in Table \ref{tab:full-dataset-results} reveal several key findings. First, all LoRA-MoE methods outperforms a single LoRA, with improvements ranging from 2.96\% (MoLORA 16x4) to 6.93\% (MoDE 16x4) in ROUGE-L scores. This demonstrates the effectiveness of MoE-based architectures in multi-task settings, allowing the model to leverage specialized experts for different tasks. 

The advantage of parameter sharing is evident in the comparison between MoLoRA 16x4 and MoLoRA-SD 16x4. By sharing the down-projection matrix, MoLoRA-SD achieves an 0.88\% improvement over MoLoRA while using only 36\% of the additional parameters, highlighting the benefit of reducing parameter redundancy.

Notably, all four models with a 16x4 configuration utilize the same number of effective parameters during inference. Among these, MoDE 16x4 achieves the highest overall performance by leveraging both shared down-projection and rank-one adapters. This showcases the effectiveness of MoDE's design in balancing parameter efficiency with the need for expressive and adaptable models in multi-task scenarios.

We further conduct a task-level analysis of all the 756 evaluation tasks and report the win rate between each pair of models in Table \ref{tab:full-dataset-results-per-task}. Each raw of the table shows the win rate of target model against models in each column. We find that all LoRA-MoE methods outperform the single LoRA baseline in around 70\%-80\% of tasks. Importantly, MoDE significantly outperforms all three baselines, passing the significance test over 50\% win rate at 0.99 confidence, demonstrating its consistent superiority across a wide range of tasks.

\paragraph{Ablation Studies}
The ablation study results (rows 6-11 in Table \ref{tab:full-dataset-results}) delve into the impact of the number of experts ($m$) and the rank of the down-projection matrix ($r$).

\textbf{\em Number of Experts (m)}: Increasing the number of experts initially improves performance (MoDE 4$\times$4 v.s. MoDE 6$\times$4), suggesting that having more experts allows for better specialization. However, further increasing the number of experts to 8 or 16 (MoDE 8$\times$4 or MoDE 16$\times$4) does not lead to any improvement in performance, suggesting diminishing returns beyond a certain point.

\textbf{\em Rank r}: For a fixed number of experts (4), increasing the rank of the LoRA matrices from 4 to 6 (MoDE 4$\times$4 vs. MoDE 4$\times$6) results in a slight performance improvement (0.6018 vs. 0.6053 ROUGE-L). This suggests that higher rank matrices can capture more nuanced information, leading to better adaptation to different tasks. Further increasing the rank of both down-projection and up-projection matrices to 8 or 16 (MoDE 4$\times$8 or MoDE 4$\times$16) leads to a decrease in performance. 


\subsection{Generalized MoDE Architecture}
To gain a deeper understanding of the impact expert rank ($p$) on model performance, we conduct two sets of experiments with the generalized MoDE architecture (Section \ref{sec:MoDE-generalization}). Following the notations $m$ and $r$, the number of experts becomes $m \times r / p$, where each expert is a rank $p$ adaptor.

\paragraph{Varying Expert Rank ($p$)} In the first set of experiments, we vary the expert rank $p$ while keeping $m$ and $r$ fixed on two $m$ and $r$ combinations (4 $\times$ 16 and 16 $\times$ 4). The results are presented in Table~\ref{tab:generalized-MoDE-increase-routers}. We observe that, with fixed $m$ and $r$, increasing the expert rank generally leads to improved performance, as indicated by higher ROUGE-L scores. This suggests that increasing the expressiveness of individual experts contributes to better overall multi-task performance.

\begin{table}[h]
\small
\centering
\begin{tabular}{ccc|cc}
\toprule
\multicolumn{3}{c|}{Model Config.}& \multirow{2}{*}{Eval} &  \multirow{2}{*}{Add. Params.} \\
$m$ & $r$ & $p$ & & \\
\midrule
4 & 16 & 16  & 58.51 & 2.71\%  \\
4 & 16 & 8  & 59.15 & 3.04\%  \\
4 & 16 & 4  & 59.56 & 3.69\%  \\
4 & 16 & 2  & 59.76 & 5.00\%  \\
4 & 16 & 1 & 59.93 & 7.62\%  \\
\midrule
16 & 4 & 4  & 58.97 & 2.71\%  \\
16 & 4 & 2  & 59.30 & 4.02\%  \\
16 & 4 & 1  & 59.91 & 6.64\%  \\
\midrule
\end{tabular}
\caption{Experiments on generalized MoDE architecture. $r$: the rank of the LoRA matrices. $p$: the rank of each experts. Number of experts: $m \times r / p$. The last column represents the total number of adapter parameters as a percentage of the total number of non-embedding parameters in Gemma-2B.}
\vskip -0.1in
\label{tab:generalized-MoDE-increase-routers}
\end{table}

\paragraph{Iso-parametric Configurations} 
In the second set of experiments, we explore iso-parametric configurations of MoDE, where the total number of added parameters remains approximately constant across different model configurations. We vary the LoRA rank $r$ (4, 8, or 16) and expert rank $p$ (from 1 to r), adjusting $m$ to maintain a consistent parameter budget. Table \ref{tab:generalized-MoDE-iso-params} presents the results of these experiments, providing insights into the trade-offs between different hyperparameter choices under a fixed resource constraint.

\begin{table}[h]
\small
\centering
\begin{tabular}{ccc|cc}
\toprule
\multicolumn{3}{c|}{Model Config.}& \multirow{2}{*}{Eval} &  \multirow{2}{*}{Add. Params.} \\
$m$ & $r$ & $p$ & & \\
\midrule
42 & 4 & 4  & 59.89 & 6.58\%  \\
27 & 4 & 2  & 60.06 & 6.55\%  \\
16 & 4 & 1  & 59.91 & 6.64\%  \\
\midrule
27 & 8 & 8  & 60.52 & 6.48\%  \\
20 & 8 & 4  & 60.74 & 6.61\%  \\
12 & 8 & 2  & 60.23 & 6.19\%  \\
7 & 8 & 1   & 59.73 & 6.18\%  \\
\midrule
15 & 16 & 16 & 60.77 & 6.55\%  \\
12 & 16 & 8 & 60.94 & 6.49\%  \\
8 & 16 & 4  & 60.77 & 6.07\%  \\
5 & 16 & 2  & 60.28 & 5.92\%  \\
3 & 16 & 1 & 59.42 & 6.04\%  \\
\bottomrule
\end{tabular}
\caption{Experiments on generalized MoDE architecture with iso-parametric constraint.}
\vskip -0.1in
\label{tab:generalized-MoDE-iso-params}
\end{table}

\begin{figure*}[t]
\centering
    \includegraphics[width=\textwidth]{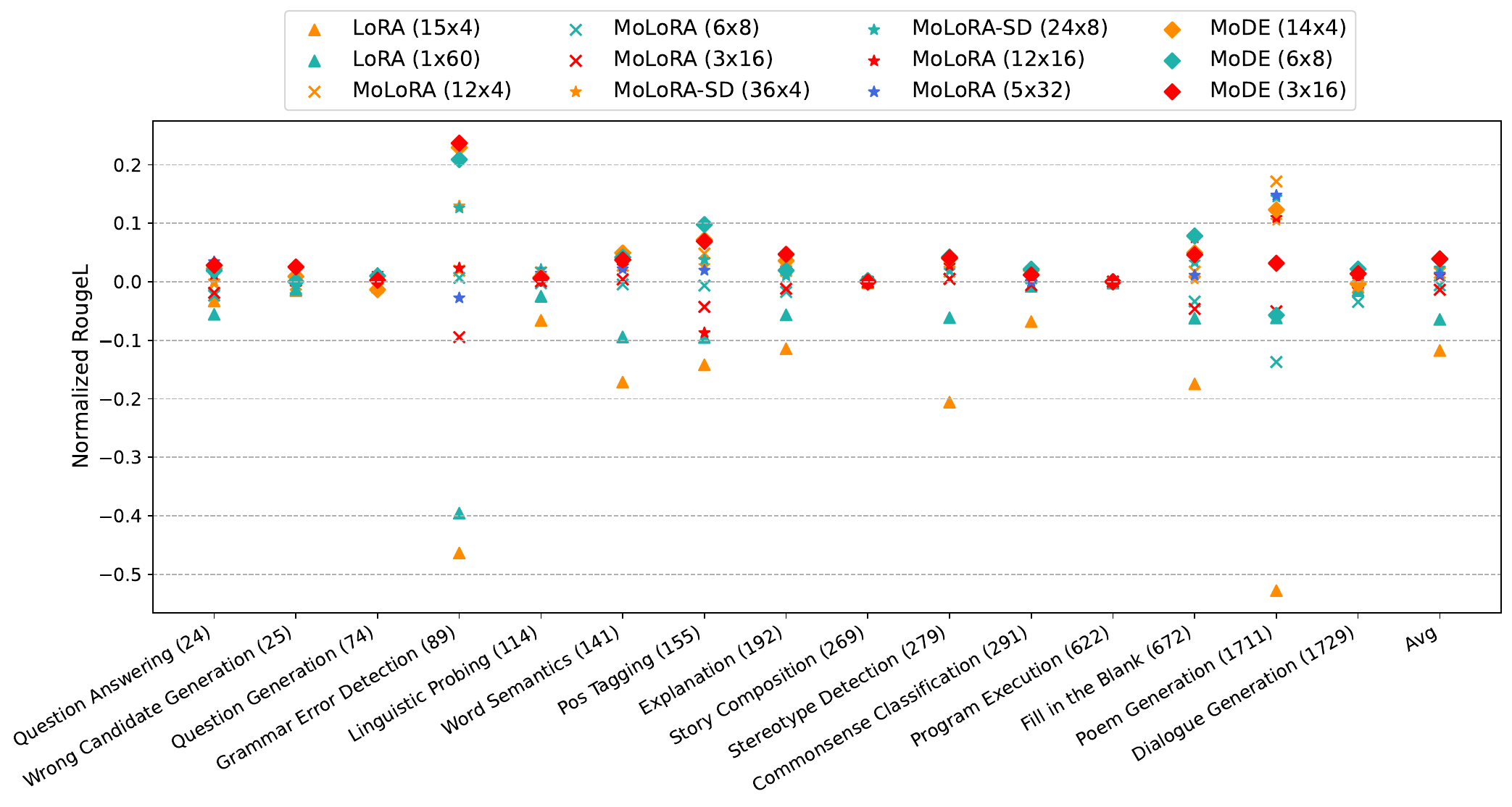}
    \caption{Performance comparison among various model configurations on 15 tasks with a fixed parameter budget.}
    \label{fig:fix-param-budget-results}
\end{figure*}

\textbf{Impact of Expert Rank ($p$)}: For a fixed LoRA rank $r$, increasing the expert rank $p$ generally improve performance, as seen when comparing configurations with the same LoRA rank but different expert ranks (e.g., 16$\times$1 v.s. 16$\times$2 v.s. 16$\times$4 v.s. 16$\times$8). This indicates that enhancing the expressiveness of individual experts with higher expert ranks contributes to better multi-task performance under the iso-parametric setting. However, gains diminish as the expert rank $p$ approaches to the LoRA rank $r$, suggesting that using $p < r$ is beneficial.

The best overall performance is achieved by the 12$\times$16$\times$8 configuration, which balances a moderate number of experts with a reasonably high LoRA rank and expert rank. This emphasize the importance of finding the optimal balance between these hyperparameters for strong multi-task performance.

\subsection{Case Study with Fixed Parameter Budget}
In real-world scenarios, there are often constraints on the number of additional parameters that can be introduced during model adaptation. To assess the effectiveness of MoDE under such constraints, we conduct a case study with a fixed parameter budget determined by the baseline models. 

We leverage the diverse subset of 15 individual tasks from the SNI dataset, each belonging to a distinct category, as described in Section \ref{subsec:dataset-and-metrics}. Our baseline model consists of 15 individual rank-4 LoRA adapters (denoted as {\em LoRA 15$\times$4}), one for each task, resulting in approximately 6 million additional trainable parameters compared to the frozen LLM backbone. This baseline establishes our parameter budget for further experimentation.

To ensure a fair comparison, we identify configurations for LoRA (trained on mixture of tasks), MoDE, MoLORA, and MoLORA-SD that introduce a similar number of parameters (approximately 6 million). We systematically explore combinations of experts ($m$) and ranks ($r$) for each method, aiming to keep the total number of additional parameters as close as possible to the baseline budget. Specifically, we experiment with ranks of 4, 8, 16, and 32, adjusting the number of experts accordingly to maintain the desired parameter count. Our notation "$m \times r$" indicates a model with m experts, each using adapters associated with a LoRA with rank r. This results in the following configurations:

\begin{itemize}
  \item {\em LoRA 1$\times$60}
  \item {\em MoRA 14$\times$4, 6$\times$8, 3$\times$16}
  \item {\em MoLORA 12$\times$4, 6$\times$8, 3$\times$16}
  \item {\em MoLORA-SD 36$\times$4, 24$\times$8, 12$\times$16, 5$\times$32}
\end{itemize}

Note that MoLORA-SD, due to its parameter efficiency from sharing the down-projection matrix, can accommodate a configuration with a higher rank (5x32) while still adhering to the budget.

\paragraph{Overall Performance} As shown in Figure \ref{fig:fix-param-budget-results}, MoDE consistently achieves comparable or superior performance to the baseline of 15 individual LoRA adapters and alternative MoE approaches (MoLORA and MoLORA-SD). This demonstrates MoDE's ability to effectively leverage its shared down-projection matrix and dyadic experts to achieve strong multi-task performance while maintaining parameter efficiency. 

Notably, MoDE achieves a substantial improvement in the overall average ROUGE-L score compared to the baseline and other MoE models. This result highlights MoDE's effectiveness in balancing parameter efficiency with the flexibility to adapt to diverse tasks.

\paragraph{Impact of Experts and Rank} While individual task performance varies across different MoDE configurations, we observe that the overall average performance across all tasks and examples remains relatively stable despite changes in the number of experts (m) and the rank (r). This suggests that MoDE's performance is robust to these hyperparameter choices, and there may not be a single "best" configuration for all scenarios. The optimal choice of experts and rank might depend on specific task characteristics, resource constraints, or desired trade-offs between model size and performance.

\paragraph{Benefit of down-projection sharing} MoLORA-SD, which shares the down-projection matrix like MoDE, generally outperforms the standard MoLORA with independent down-projection matrices. This highlights the importance of reducing parameter redundancy and promoting knowledge sharing across tasks through a shared down-projection matrix.

\paragraph{Benefit of Mixture of Dyadic Experts} Comparing MoDE and MoLORA-SD configurations, we observe that MoDE often achieves better performance. This suggests that the dyadic experts in MoDE contribute to its superior expressivity and adaptability compared to simply sharing the down-projection matrix.

These findings demonstrates that MoDE can effectively leverage a fixed parameter budget to achieve strong multi-task performance. Its shared down-projection matrix and mixtures of dyadic experts enable a balance between parameter efficiency and expressive power, making it a promising approach for deploying multi-task LLMs in resource-constrained environments.




\section{Conclusion and Future Work}
\label{sec:conclusion}

In this paper, we introduce MoDE (Mixture of Dyadic Experts), a novel parameter-efficient fine-tuning method for multi-task adaptation of large language models. MoDE addresses the limitations of existing LoRA-based MoE architectures by sharing the down-projection matrix across experts to remove parameter redundancy. The key innovation of MoDE lies in its use of rank-one adapters, combined with a sophisticated routing mechanism that allows for nuanced and flexible combinations of these adapters. This design fosters knowledge sharing, reduces redundancy, and increases expressive power, enabling the model to effectively capture the unique characteristics of each task.

Our experiments on the Supernatural Instructions benchmark demonstrate that MoDE consistently outperforms state-of-the-art multi-task PEFT methods, achieving superior performance with comparable parameter efficiency. This highlight MoDE's potential as an efficient and effective solution for multi-task LLM adaptation, particularly in resource-constrained environments.


Future work will explore top-k routing strategies to further enhance MoDE's efficiency and adaptability. We also plan to analyze the routing behavior to identify task-specific patterns, and assess MoDE's generalization capabilities to unseen tasks. Additionally, we aim to evaluate MoDE on larger models and alternative PEFT techniques.

\clearpage
\section{Limitations}

While our proposed MoDE architecture demonstrates promising results in multi-task LLM adaptation, there are several limitations that warrant further investigation.

\paragraph{Routing Strategy} The current MoDE implementation utilizes a relatively simple routing mechanism based on a softmax function. While effective in our experiments, exploring more sophisticated routing strategies that incorporate task relationships or input-specific features could potentially further improve performance.

\paragraph{Hyperparameter Sensitivity} The optimal number of experts and rank of the LoRA matrices can vary depending on the specific task distribution and available resources. While our ablation study provides some insights, a more comprehensive exploration of hyperparameter sensitivity could help identify optimal configurations for different scenarios.

\paragraph{Computational Overhead} While MoDE significantly reduces parameter count compared to traditional LoRA-MoE, the routing mechanism introduces additional computational overhead during inference. This overhead could become a bottleneck in real-time applications with strict latency requirements. Investigating ways to optimize the routing process or reduce its computational cost would be beneficial.

\paragraph{Evaluation Benchmark} Our evaluation primarily focuses on the Supernatural Instructions benchmark. While this dataset covers a wide range of tasks, it may not fully represent the diversity of real-world applications. Evaluating MoDE on other multi-task benchmarks or in specific domains could further validate its effectiveness and generalizability.
Addressing these limitations could lead to even more efficient and adaptable multi-task LLM architectures, further expanding the potential of parameter-efficient fine-tuning for a wider range of applications.

\bibliography{references}

\begin{thebibliography}{26}
\providecommand{\natexlab}[1]{#1}

\bibitem[{Brown et~al.(2020)Brown, Mann, Ryder, Subbiah, Kaplan, Dhariwal,
  Neelakantan, Shyam, Sastry, Askell et~al.}]{brown2020language}
Tom Brown, Benjamin Mann, Nick Ryder, Melanie Subbiah, Jared~D Kaplan, Prafulla
  Dhariwal, Arvind Neelakantan, Pranav Shyam, Girish Sastry, Amanda Askell,
  et~al. 2020.
\newblock Language models are few-shot learners.
\newblock \emph{Advances in neural information processing systems},
  33:1877--1901.

\bibitem[{Caruana(1997)}]{caruana1997multitask}
Rich Caruana. 1997.
\newblock Multitask learning.
\newblock \emph{Machine learning}, 28:41--75.

\bibitem[{Fedus et~al.(2022)Fedus, Zoph, and
  Shazeer}]{fedus2022switchtransformersscalingtrillion}
William Fedus, Barret Zoph, and Noam Shazeer. 2022.
\newblock \href {https://arxiv.org/abs/2101.03961} {Switch transformers:
  Scaling to trillion parameter models with simple and efficient sparsity}.
\newblock \emph{Preprint}, arXiv:2101.03961.

\bibitem[{Feng et~al.(2024)Feng, Hao, Zhang, Han, and Wang}]{feng2024mixture}
Wenfeng Feng, Chuzhan Hao, Yuewei Zhang, Yu~Han, and Hao Wang. 2024.
\newblock Mixture-of-loras: An efficient multitask tuning for large language
  models.
\newblock \emph{arXiv preprint arXiv:2403.03432}.

\bibitem[{Hu et~al.(2021)Hu, Shen, Wallis, Allen-Zhu, Li, Wang, Wang, and
  Chen}]{hu2021lora}
Edward~J Hu, Yelong Shen, Phillip Wallis, Zeyuan Allen-Zhu, Yuanzhi Li, Shean
  Wang, Lu~Wang, and Weizhu Chen. 2021.
\newblock Lora: Low-rank adaptation of large language models.
\newblock \emph{arXiv preprint arXiv:2106.09685}.

\bibitem[{Huang et~al.(2023)Huang, Liu, Lin, Pang, Du, and
  Lin}]{huang2023lorahub}
Chengsong Huang, Qian Liu, Bill~Yuchen Lin, Tianyu Pang, Chao Du, and Min Lin.
  2023.
\newblock Lorahub: Efficient cross-task generalization via dynamic lora
  composition.
\newblock \emph{arXiv preprint arXiv:2307.13269}.

\bibitem[{Jiang et~al.(2024)Jiang, Sablayrolles, Roux, Mensch, Savary, Bamford,
  Chaplot, de~las Casas, Hanna, Bressand, Lengyel, Bour, Lample, Lavaud,
  Saulnier, Lachaux, Stock, Subramanian, Yang, Antoniak, Scao, Gervet, Lavril,
  Wang, Lacroix, and Sayed}]{jiang2024mixtralexperts}
Albert~Q. Jiang, Alexandre Sablayrolles, Antoine Roux, Arthur Mensch, Blanche
  Savary, Chris Bamford, Devendra~Singh Chaplot, Diego de~las Casas, Emma~Bou
  Hanna, Florian Bressand, Gianna Lengyel, Guillaume Bour, Guillaume Lample,
  Lélio~Renard Lavaud, Lucile Saulnier, Marie-Anne Lachaux, Pierre Stock,
  Sandeep Subramanian, Sophia Yang, Szymon Antoniak, Teven~Le Scao, Théophile
  Gervet, Thibaut Lavril, Thomas Wang, Timothée Lacroix, and William~El Sayed.
  2024.
\newblock \href {https://arxiv.org/abs/2401.04088} {Mixtral of experts}.
\newblock \emph{Preprint}, arXiv:2401.04088.

\bibitem[{Li et~al.(2024)Li, Ma, Wang, Cheng, Duan, Zuo, Yang, and
  Tang}]{li2024mixlora}
Dengchun Li, Yingzi Ma, Naizheng Wang, Zhiyuan Cheng, Lei Duan, Jie Zuo, Cal
  Yang, and Mingjie Tang. 2024.
\newblock Mixlora: Enhancing large language models fine-tuning with lora based
  mixture of experts.
\newblock \emph{arXiv preprint arXiv:2404.15159}.

\bibitem[{Li et~al.(2022)Li, Gururangan, Dettmers, Lewis, Althoff, Smith, and
  Zettlemoyer}]{li2022branchtrainmergeembarrassinglyparalleltraining}
Margaret Li, Suchin Gururangan, Tim Dettmers, Mike Lewis, Tim Althoff, Noah~A.
  Smith, and Luke Zettlemoyer. 2022.
\newblock \href {https://arxiv.org/abs/2208.03306} {Branch-train-merge:
  Embarrassingly parallel training of expert language models}.
\newblock \emph{Preprint}, arXiv:2208.03306.

\bibitem[{Lin et~al.(2024)Lin, Wang, Clark, Lu, Zhu, Whitehouse, and
  Yu}]{lin2024multitask}
Chu-Cheng Lin, Xinyi Wang, Jonathan~H Clark, Han Lu, Yun Zhu, Chenxi
  Whitehouse, and Hongkun Yu. 2024.
\newblock Multitask multilingual model adaptation with featurized low-rank
  mixtures.
\newblock \emph{arXiv preprint arXiv:2402.17934}.

\bibitem[{Liu et~al.(2023)Liu, Wu, Zhao, Zhu, Xu, Tian, and
  Zheng}]{liu2023moelora}
Qidong Liu, Xian Wu, Xiangyu Zhao, Yuanshao Zhu, Derong Xu, Feng Tian, and
  Yefeng Zheng. 2023.
\newblock Moelora: An moe-based parameter efficient fine-tuning method for
  multi-task medical applications.
\newblock \emph{arXiv preprint arXiv:2310.18339}.

\bibitem[{Liu et~al.(2024)Liu, Zhang, Yang, Keutzer, Du, Du, and
  Zhang}]{liu2024intuitionaware}
Yijiang Liu, Rongyu Zhang, Huanrui Yang, Kurt Keutzer, Yuan Du, Li~Du, and
  Shanghang Zhang. 2024.
\newblock \href {https://arxiv.org/abs/2404.08985} {Intuition-aware
  mixture-of-rank-1-experts for parameter efficient finetuning}.
\newblock \emph{Preprint}, arXiv:2404.08985.

\bibitem[{OpenAI and et~al.(2024)}]{openai2024gpt4technicalreport}
OpenAI and Josh~Achiam et~al. 2024.
\newblock \href {https://arxiv.org/abs/2303.08774} {Gpt-4 technical report}.
\newblock \emph{Preprint}, arXiv:2303.08774.

\bibitem[{Ostapenko et~al.(2024)Ostapenko, Su, Ponti, Charlin, Roux, Pereira,
  Caccia, and Sordoni}]{ostapenko2024towards}
Oleksiy Ostapenko, Zhan Su, Edoardo~Maria Ponti, Laurent Charlin, Nicolas~Le
  Roux, Matheus Pereira, Lucas Caccia, and Alessandro Sordoni. 2024.
\newblock Towards modular llms by building and reusing a library of loras.
\newblock \emph{arXiv preprint arXiv:2405.11157}.

\bibitem[{Ruder(2017)}]{ruder2017overview}
Sebastian Ruder. 2017.
\newblock An overview of multi-task learning in deep neural networks.
\newblock \emph{arXiv preprint arXiv:1706.05098}.

\bibitem[{Shah et~al.(2023)Shah, Ruiz, Cole, Lu, Lazebnik, Li, and
  Jampani}]{shah2023ziplora}
Viraj Shah, Nataniel Ruiz, Forrester Cole, Erika Lu, Svetlana Lazebnik,
  Yuanzhen Li, and Varun Jampani. 2023.
\newblock Ziplora: Any subject in any style by effectively merging loras.
\newblock \emph{arXiv preprint arXiv:2311.13600}.

\bibitem[{Shazeer et~al.(2017)Shazeer, Mirhoseini, Maziarz, Davis, Le, Hinton,
  and Dean}]{shazeer2017outrageously}
Noam Shazeer, Azalia Mirhoseini, Krzysztof Maziarz, Andy Davis, Quoc Le,
  Geoffrey Hinton, and Jeff Dean. 2017.
\newblock Outrageously large neural networks: The sparsely-gated
  mixture-of-experts layer.
\newblock \emph{arXiv preprint arXiv:1701.06538}.

\bibitem[{Shazeer and Stern(2018)}]{shazeer2018adafactor}
Noam Shazeer and Mitchell Stern. 2018.
\newblock Adafactor: Adaptive learning rates with sublinear memory cost.
\newblock In \emph{International Conference on Machine Learning}, pages
  4596--4604. PMLR.

\bibitem[{Sukhbaatar et~al.(2024)Sukhbaatar, Golovneva, Sharma, Xu, Lin,
  Rozière, Kahn, Li, tau Yih, Weston, and
  Li}]{sukhbaatar2024branchtrainmixmixingexpertllms}
Sainbayar Sukhbaatar, Olga Golovneva, Vasu Sharma, Hu~Xu, Xi~Victoria Lin,
  Baptiste Rozière, Jacob Kahn, Daniel Li, Wen tau Yih, Jason Weston, and Xian
  Li. 2024.
\newblock \href {https://arxiv.org/abs/2403.07816} {Branch-train-mix: Mixing
  expert llms into a mixture-of-experts llm}.
\newblock \emph{Preprint}, arXiv:2403.07816.

\bibitem[{Team et~al.(2023)Team, Anil, Borgeaud, Wu, Alayrac, Yu, Soricut,
  Schalkwyk, Dai, Hauth et~al.}]{team2023gemini}
Gemini Team, Rohan Anil, Sebastian Borgeaud, Yonghui Wu, Jean-Baptiste Alayrac,
  Jiahui Yu, Radu Soricut, Johan Schalkwyk, Andrew~M Dai, Anja Hauth, et~al.
  2023.
\newblock Gemini: a family of highly capable multimodal models.
\newblock \emph{arXiv preprint arXiv:2312.11805}.

\bibitem[{Team et~al.(2024)Team, Mesnard, Hardin, Dadashi, Bhupatiraju, Pathak,
  Sifre, Rivi{\`e}re, Kale, Love et~al.}]{team2024gemma}
Gemma Team, Thomas Mesnard, Cassidy Hardin, Robert Dadashi, Surya Bhupatiraju,
  Shreya Pathak, Laurent Sifre, Morgane Rivi{\`e}re, Mihir~Sanjay Kale,
  Juliette Love, et~al. 2024.
\newblock Gemma: Open models based on gemini research and technology.
\newblock \emph{arXiv preprint arXiv:2403.08295}.

\bibitem[{Wang et~al.(2022{\natexlab{a}})Wang, Agarwal, Mukherjee, Liu, Gao,
  Awadallah, and Gao}]{wang2022adamix}
Yaqing Wang, Sahaj Agarwal, Subhabrata Mukherjee, Xiaodong Liu, Jing Gao,
  Ahmed~Hassan Awadallah, and Jianfeng Gao. 2022{\natexlab{a}}.
\newblock Adamix: Mixture-of-adaptations for parameter-efficient model tuning.
\newblock \emph{arXiv preprint arXiv:2205.12410}.

\bibitem[{Wang et~al.(2022{\natexlab{b}})Wang, Mishra, Alipoormolabashi, Kordi,
  Mirzaei, Arunkumar, Ashok, Dhanasekaran, Naik, Stap et~al.}]{wang-2022-sni}
Yizhong Wang, Swaroop Mishra, Pegah Alipoormolabashi, Yeganeh Kordi, Amirreza
  Mirzaei, Anjana Arunkumar, Arjun Ashok, Arut~Selvan Dhanasekaran, Atharva
  Naik, David Stap, et~al. 2022{\natexlab{b}}.
\newblock Super-naturalinstructions:generalization via declarative instructions
  on 1600+ tasks.
\newblock In \emph{EMNLP}.

\bibitem[{Wu et~al.(2024)Wu, Huang, and Wei}]{wu2024mixture}
Xun Wu, Shaohan Huang, and Furu Wei. 2024.
\newblock \href {https://openreview.net/forum?id=uWvKBCYh4S} {Mixture of lora
  experts}.

\bibitem[{Zadouri et~al.(2023)Zadouri, {\"U}st{\"u}n, Ahmadian, Ermi{\c{s}},
  Locatelli, and Hooker}]{zadouri2023pushing}
Ted Zadouri, Ahmet {\"U}st{\"u}n, Arash Ahmadian, Beyza Ermi{\c{s}}, Acyr
  Locatelli, and Sara Hooker. 2023.
\newblock Pushing mixture of experts to the limit: Extremely parameter
  efficient moe for instruction tuning.
\newblock \emph{arXiv preprint arXiv:2309.05444}.

\bibitem[{Zhu et~al.(2023)Zhu, Wichers, Lin, Wang, Chen, Shu, Lu, Liu, Luo,
  Chen et~al.}]{zhu2023sira}
Yun Zhu, Nevan Wichers, Chu-Cheng Lin, Xinyi Wang, Tianlong Chen, Lei Shu, Han
  Lu, Canoee Liu, Liangchen Luo, Jindong Chen, et~al. 2023.
\newblock Sira: Sparse mixture of low rank adaptation.
\newblock \emph{arXiv preprint arXiv:2311.09179}.

\end{thebibliography}

\appendix

\clearpage
\onecolumn

\section{Detailed Results for Case Study with Fixed Parameter Budge}
Table \ref{tab:fix-param-budget-results} presents the detailed model performance with 12 model configures on the 15 tasks.

\begin{table*}[h]
\small
\centering
\begin{tabular}{l|cc|ccc|cccc|ccc} 
\toprule
\multirow{2}{*}{Task ID} & \multicolumn{2}{c|}{LoRA} & \multicolumn{3}{c|}{MoLORA} & \multicolumn{4}{c|}{MoLORA-SD} & \multicolumn{3}{c}{MoDE} \\
\cline{2-13}
& 15$\times$4 & 1$\times$60 & 12$\times$4 & 6$\times$8 & 3$\times$16  & 36$\times$4  & 24$\times$8 & 12$\times$6 & 5$\times$32 & 14$\times$4 & 6$\times$8 & 3$\times$16 \\ 
\midrule
task24 & 25.91 & 25.30 & 26.67 & 26.16 & 26.29 & 26.81 & 27.26 & 27.12 & \textcolor{blue}{\textbf{27.71}} & 27.40 & 27.31 & \underline{\textcolor[rgb]{0,0.502,0}{\bf 27.55}} \\
task25 & 41.27 & 41.35 & 41.7 & 41.43 & 41.99 & 41.91 & 41.73 & 41.92 & \underline{\textcolor[rgb]{0,0.502,0}{\bf 42.36}} & 42.29 & 41.92 & \textcolor{blue}{\textbf{42.97}} \\
task74 & 42.73 & 42.53 & 43.12 & 42.86 & 42.64 & 42.67 & 42.88 & \textcolor{blue}{\textbf{43.53}} & 43.30 & 42.34 & \underline{\textcolor[rgb]{0,0.502,0}{\bf 43.35}} & 43.03 \\
task89 & 33.90  & 38.21 & 64.41 & 63.64 & 57.24 & 71.42 & 71.19 & 64.71 & 61.48 & \underline{\textcolor[rgb]{0,0.502,0}{\bf 77.73}} & 76.43 & \textcolor{blue}{\textbf{78.20}} \\
task114 & 84.15 & 87.85 & \underline{\textcolor[rgb]{0,0.502,0}{\bf 91.54}} & 89.85 & 90.15 & 91.08 & \textcolor{blue}{\textbf{92.00}} & 90.92 & 91.38 & 90.77 & 90.62 & 90.62 \\
task141 & 75.79 & 82.87 & 94.00 & 91.08 & 91.85 & 94.15 & 94.31 & 93.95 & 93.49 & \textcolor{blue}{\textbf{96.00}} & \underline{\textcolor[rgb]{0,0.502,0}{\bf 95.33}} & 94.92 \\
task155 & 51.09 & 53.88 & 62.42 & 59.16 & 56.99 & 61.18 & 61.80 & 54.35 & 60.71 & \underline{\textcolor[rgb]{0,0.502,0}{\bf 63.82}} & \textcolor{blue}{\textbf{65.37}} & 63.66 \\
task192 & 72.47 & 77.21 & 83.35 & 80.41 & 80.86 & 82.99 & 82.68 & 84.47 & 83.61 & \underline{\textcolor[rgb]{0,0.502,0}{\bf 84.78}} & 83.42 & \textcolor{blue}{\textbf{85.69}} \\
task269 & 75.40  & 75.51 & 75.46 & 75.56 & 75.50  & 75.61 & 75.52 & 75.33 & \textcolor{blue}{\textbf{75.72}} & 75.59 & \underline{\textcolor[rgb]{0,0.502,0}{\bf 75.71}} & 75.52 \\
task279 & 69.90  & 82.59 & 89.52 & 88.39 & 88.44 & 90.60  & 89.68 & 90.45 & 91.42 & 91.53 & \textcolor{blue}{\textbf{91.73}} & \underline{\textcolor[rgb]{0,0.502,0}{\bf 91.58}} \\
task291 & 79.97 & 85.14 & 86.64 & 86.31 & 85.31 & 86.31 & 86.31 & 85.81 & 85.81 & \underline{\textcolor[rgb]{0,0.502,0}{\bf 87.48}} & \textcolor{blue}{\textbf{87.65}} & 86.81 \\
task622 & 99.95 & 99.70 & \textcolor{blue}{\textbf{99.97}} & 99.93 &\underline{\textcolor[rgb]{0,0.502,0}{\bf 99.96}} & \underline{\textcolor[rgb]{0,0.502,0}{\bf 99.96}} & 99.93 & 99.93 & 99.94 & \underline{\textcolor[rgb]{0,0.502,0}{\bf 99.96}} & 99.94 & 99.93 \\
task672 & 39.69 & 45.08 & 48.92 & 46.46 & 45.85 & 48.31 & 49.69 & \underline{\textcolor[rgb]{0,0.502,0}{\bf 51.69}}  & 48.62 & 50.46 & \textcolor{blue}{\textbf{51.85}} & 50.31 \\
task1711 & 4.19 & 8.33 & \textcolor{blue}{\textbf{10.40}} & 7.66 & 8.43 & 9.81 & 10.16 & 9.85 & \underline{\textcolor[rgb]{0,0.502,0}{\bf 10.19}} & 9.97 & 8.37 & 9.16 \\
task1729 & 17.21 & 17.11 & 17.43 & 16.77 & 17.34 & 17.25 & 17.35 & 17.60 & \underline{\textcolor[rgb]{0,0.502,0}{\bf 17.68}} & 17.31 & \textcolor{blue}{\textbf{17.75}} & 17.61 \\
\midrule
Overall & 53.83 & 57.07 & 61.93 & 60.63 & 60.16 & 62.25 & 62.42 & 61.66 & 61.79 & \textcolor{blue}{\textbf{63.39}} & 63.37 & \textcolor{blue}{\textbf{63.39}} \\
\bottomrule
\end{tabular}
\caption{Performance comparison among various model configurations on 15 tasks with a fixed parameter budget. The scores in blue and green correspond to the highest and second-highest scores for the corresponding task.}
\label{tab:fix-param-budget-results}
\end{table*}

\section{PCA Clustering of LoRA Matrices }
\label{sec:pca_all_matrices}

\begin{figure*}[t]
\centering
    \includegraphics[width=0.70\textwidth]{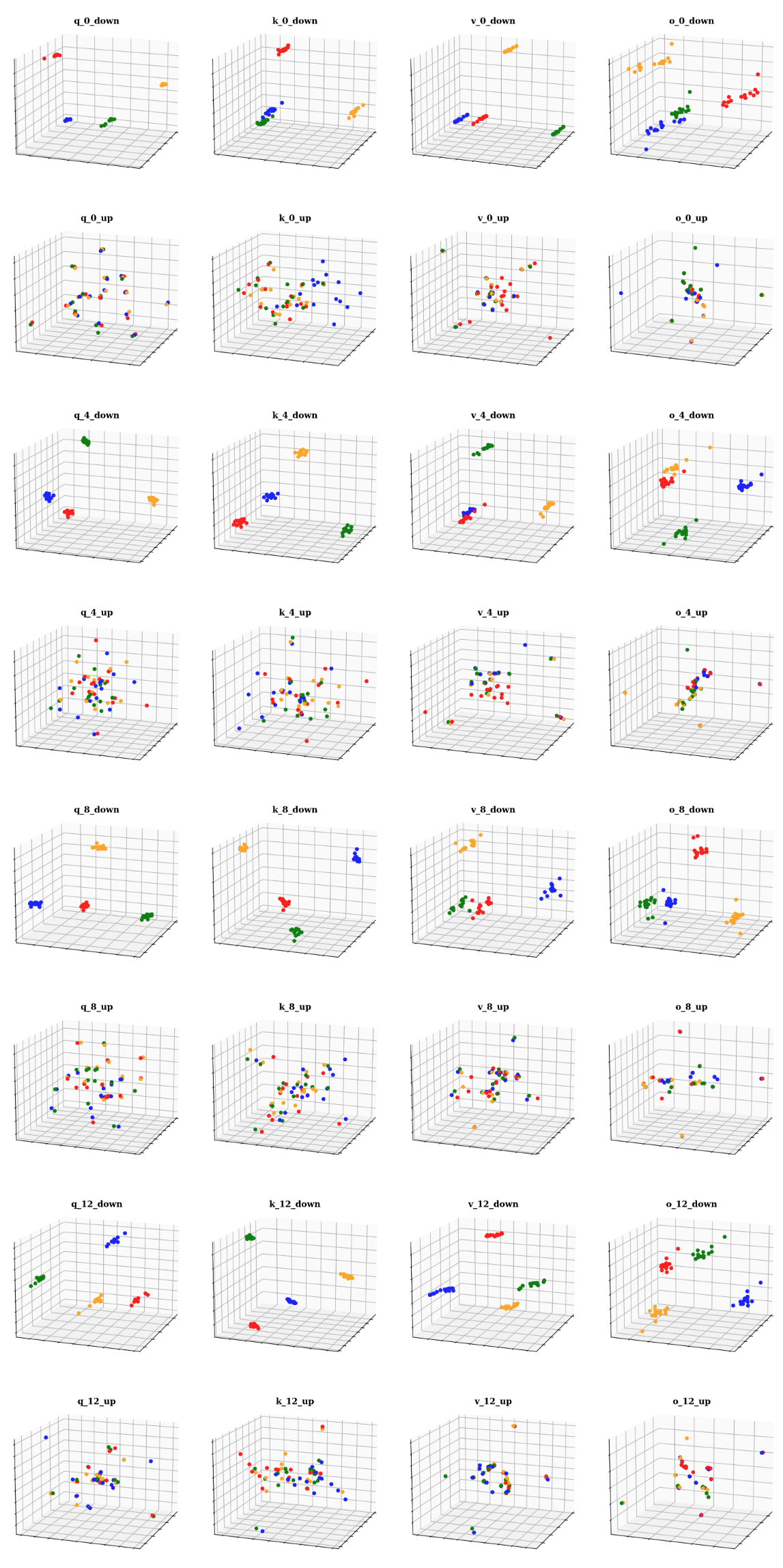}
    \vskip -0.1in
    \caption{Scatter plots after applying Principal Component Analysis (PCA) on all  LoRA projection matrices, i.e. query, key, value, and output, sliced along the rank dimension.The clear clustering of down-projection vectors suggests the presence of shared representations across tasks, motivating the design of the MoDE architecture.}
    \vskip -0.1in
    \label{fig:PCA_plots_all}
\end{figure*}

Scatter plots after applying Principal Component Analysis (PCA) on all of the LoRA projection matrices at layer 6 and 12 are shown in Figure \ref{fig:PCA_plots_all}. The distinct grouping of down-projection vectors indicates common representations across tasks, providing the inspiration for the MoDE architecture.

\end{document}